\documentclass[nohyperref]{article}

\usepackage{microtype}
\usepackage{graphicx}
\usepackage{subfigure}
\usepackage{booktabs} 

\usepackage{hyperref}

\usepackage[accepted]{icml2023}

\usepackage{amsmath}
\usepackage{amssymb}
\usepackage{mathtools}
\usepackage{amsthm}

\usepackage[capitalize,noabbrev]{cleveref}

\theoremstyle{plain}

\theoremstyle{definition}

\theoremstyle{remark}

\usepackage[textsize=tiny]{todonotes}

\icmltitlerunning{Spuriosity Rankings for Free: A Simple Framework for Last Layer Retraining Based on Object Detection}

\begin{document}

\twocolumn[
\icmltitle{Spuriosity Rankings for Free: \\ A Simple Framework for Last Layer Retraining Based on Object Detection}



\icmlsetsymbol{equal}{*}

\begin{icmlauthorlist}
\icmlauthor{Mohammad Azizmalayeri}{equal,yyy}
\icmlauthor{Reza Abbasi}{equal,yyy}
\icmlauthor{Amir Hosein Haji Mohammad rezaie}{equal,yyy}
\icmlauthor{Reihaneh Zohrabi}{equal,yyy}
\icmlauthor{Mahdi Amiri}{equal,yyy}
\icmlauthor{Mohammad Taghi Manzuri}{yyy}
\icmlauthor{Mohammad Hossein Rohban}{yyy}
\end{icmlauthorlist}

\icmlaffiliation{yyy}{Department of Computer Engineering, Sharif University of Technology, Tehran, Iran}

\icmlcorrespondingauthor{Mohammad Azizmalayeri}{m.azizmalayeri@sharif.edu}
\icmlcorrespondingauthor{Reza Abbasi}{reza.abbasi@sharif.edu}

\icmlkeywords{Machine Learning, ICML}

\vskip 0.3in
]



\printAffiliationsAndNotice{\icmlEqualContribution} 

\begin{figure*}[t]
  \centering
  \includegraphics[width=\textwidth]{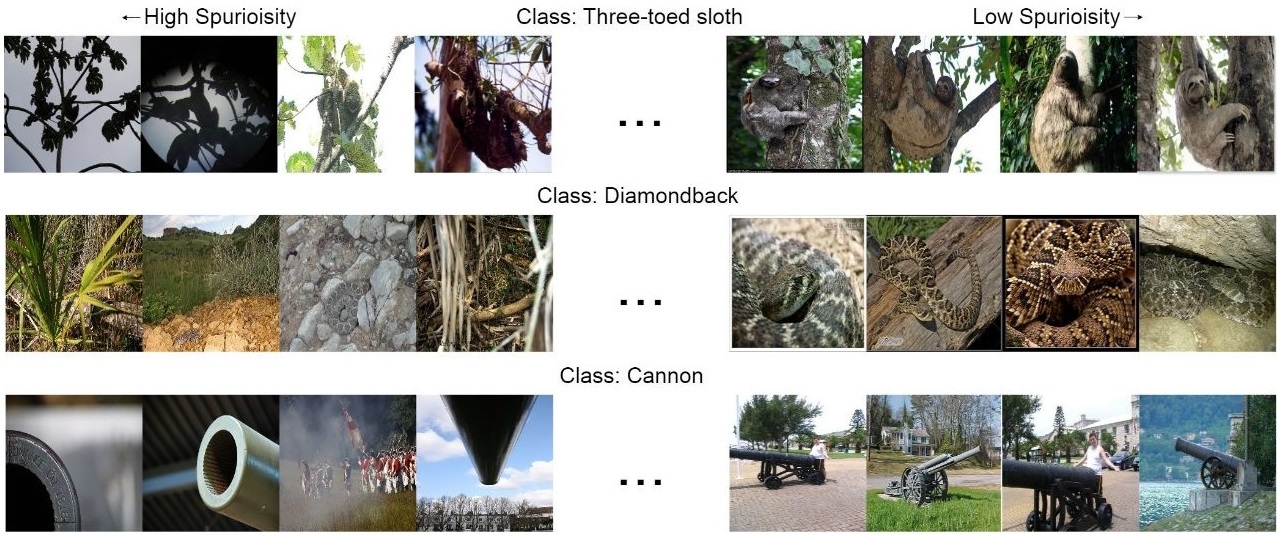} 
  \vskip -0.05in
  \caption{
  Spuriosity ranking reveals core object detectability and the presence of spurious cues, distinguishing between clear representations such as a recognizable three-toed sloth, Diamondback, and Cannon (low spurious cues - right) and more ambiguous instances featuring spurious cues like trees with a less prominent three-toed sloth, Diamondbacks in hard-to-distinguish locations, and Cannon barrel (high spurious cues - left).
  }
  \label{fig1} 
  \vskip -0.05in
\end{figure*}

\begin{abstract}

Deep neural networks have exhibited remarkable performance in various domains. However, the reliance of these models on spurious features has raised concerns about their reliability. A promising solution to this problem is last-layer retraining, which involves retraining the linear classifier head on a small subset of data without spurious cues. Nevertheless, selecting this subset requires human supervision, which reduces its scalability. Moreover, spurious cues may still exist in the selected subset. As a solution to this problem, we propose a novel ranking framework that leverages an open vocabulary object detection technique to identify images without spurious cues. More specifically, we use the object detector as a measure to score the presence of the target object in the images. Next, the images are sorted based on this score, and the last-layer of the model is retrained on a subset of the data with the highest scores. Our experiments on the ImageNet-1k dataset demonstrate the effectiveness of this ranking framework in sorting images based on spuriousness and using them for last-layer retraining.


\end{abstract}

\section{Introduction}


A prominent issue that has recently gained substantial attention is the dependency of deep models on shortcuts 
\cite{shortcut}.
An illustrative example of such shortcuts is spurious correlation which is statistically informative heuristics for predicting the labels of most examples in the training data. However, they are not actually relevant to the true labeling function, and do not necessarily have a causal relationship with the class label \cite{lastlayer}. A model that captures these spurious correlations may have high average accuracy on a given test set, but encounters significant accuracy drop on data subsets without these correlations \cite{IRM,DRO,alcorn2019strike}. 
Theoretical arguments suggest that a bias towards model simplicity contributes to the reliance on spurious features \cite{shah2020pitfalls}. 

A wide range of solutions have been proposed to address the model robustness against spurious correlations. One approach involves optimizing for the worst case group accuracy, which necessitates the annotations indicating the respective group affiliation of each training example \cite{DRO,zhang2021coping, liu2021just, hu2018does}. Additionally, alternative strategies involve leveraging causality \cite{aubin2021linear, seo2022informationtheoretic}, and learning invariant latent spaces \cite{IRM, robey2021modelbased}. Recent researches have also highlighted the effectiveness of fine-tuning the last layer of neural networks on samples without spurious correlations to the label, to mitigate the model sensitivity to spurious cues \cite{lastlayer, rosenfeld2022domainadjusted, lee2023surgical}. 
However, they often rely on human supervision to determine the presence of spurious features, which may limit their applicability, particularly for large datasets like ImageNet \cite{5206848}. For example, in \cite{lastlayer}, human supervision is necessary to re-weight the spurious features through retraining the last linear layer on a small dataset without spurious correlations between the background and foreground.

To overcome this limitation,
\cite{singla2022salient, singla2022core} have introduced an approach that utilizes the neurons of robust models as detectors for visual attributes, enabling  identification of the spurious features at a large scale while minimizing the need for human supervision. Moreover, \cite{neuhaus2022spurious} have proposed an automated pipeline for detecting spurious features using class-wise neural PCA components, 
which still requires human supervision. Building upon the concept of robust neural features, \cite{moayeri2022spuriosity} introduced a novel ranking framework that assesses the prominence of spurious cues within image classes. In order to reduce the impact of spurious features, their approach involves training linear heads on carefully selected subsets of data determined by the ranking. 

Despite these notable attempts to reduce the reliance on human supervision, the complete elimination of such supervision remains challenging. The mentioned approaches still require human involvement in annotating feature-class dependencies, indicating that previous works may face limitations when employing a human-in-the-loop strategy. Therefore, it is important to acknowledge that the need for human involvement in these methods introduces scalability concerns. To overcome these challenges and further enhance the autonomy of the system, we propose a novel method for eliminating the need for human supervision in the detection and mitigation of the effect of spurious features. 

In this work, we introduce a novel approach to identify the core and spurious features within classes, utilizing an open vocabulary object detection technique \cite{objectdetection}. By examining the images with the highest and lowest spuriosity ranks, as illustrated in \cref{fig1}, we are able to detect images with the high spurious cues and minority subgroups. In our framework, images that clearly have the object, such as a clear identifiable three-toed sloth, receive high scores and are ranked as low spurious. Conversely, images where the spurious cues are present, and the object could not be easily detected, like trees with a less evident three-toed sloth, are assigned low scores and ranked as highly spurious. This approach allows for precise differentiation between images based on their level of spuriousness and object detectability, enabling us to rank images based on their spuriosity without any reliance on human supervision.

\section{Proposed Sorting Approach}

\begin{figure*}[t]
  \centering
  \includegraphics[width=\textwidth]{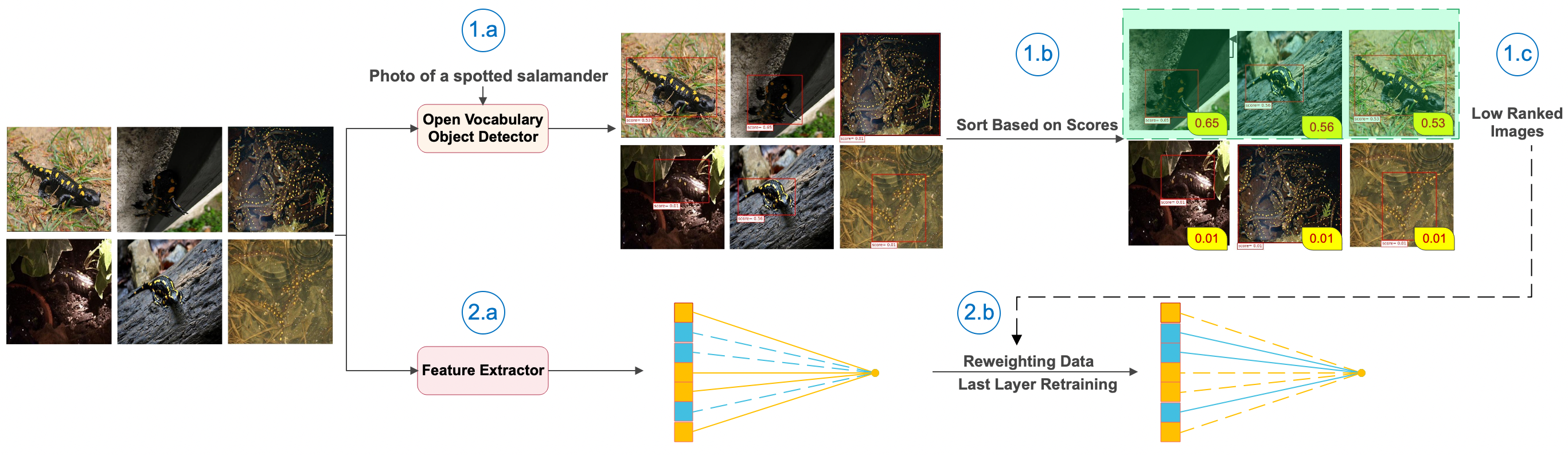} 
  \vskip -0.05in
  \caption{Schematic representation of our proposed 2-step approach.
1: a) utilizing an open-vocabulary object detector, b) sorting based on detection scores, and c) selecting low-ranked images.
2: a) training a model on the entire dataset and b) retraining the last layer.}
  \label{fig2} 
  \vskip -0.05in
\end{figure*}


One of the most effective methods for reducing the reliance of models on spurious features is the last layer re-training approach \cite{lastlayer}. To leverage this approach effectively, images within a dataset should be ranked based on their spuriosity in order to select a subset that contains low spurious cues for retraining. For large datasets, the existing ranking method \cite{moayeri2022spuriosity} requires human supervision, which may not be feasible.  Hence, we propose a method for determining feature-class dependencies without requiring human supervision. Our ranking method is based on large-scale pre-trained models that can generalize to out-of-distribution (OOD) samples.


Recent advances have applied language capabilities to object detection \cite{gu2022openvocabulary, kamath2021mdetr, objectdetection}, integrating image and text in a shared representation space\cite{NIPS2013_7cce53cf}. OWL-ViT \cite{objectdetection}, a cutting-edge method, combines image and text encoders using contrastive learning on a large dataset. Inspired by the CLIP model, OWL-ViT exhibits remarkable generalization and detects objects in open-vocabulary environments. Its generalization capability can help us detect high-spurious samples.
OWL-ViT architecture consists of a Vision Transformer \cite{dosovitskiy2021image} for image encoding and a  Transformer architecture for text encoding. The model image and text encoders are trained on a large-scale image-text data using contrastive learning. Then, detection heads are added to the model, and it is trained on a medium-sized detection data. OWL-ViT modifies the image encoder for object detection by projecting token representations and using a small MLP to determine the box coordinates. Predicted bounding boxes are ranked based on the confidence scores. These scores are calculated by transforming logits from a fully connected layer with a sigmoid function to obtain class probability scores. Aggregating the scores across classes produces a single score for each predicted bounding box.
Notably, this score serves as a valuable metric for assessing the level of spurious features within an image and can effectively rank images within their respective classes. Specifically, higher scores correspond to lower rankings of spuriosity, indicating a more clear and more identifiable object in the image.

On this basis, as shown in \cref{fig2}, our proposed method utilizes an open-vocabulary object detector to detect and classify objects in images. First, the model assigns detection scores to the objects based on its confidence. The images are then sorted based on these scores to refine the results and prioritize them according to their spuriosity.

To mitigate the impact of spurious correlations learned by the model during the initial detection process, the last layer of the model is retrained. This technique allows for fine-tuning the parameters in the last layer, enhancing the model ability to recognize meaningful features while disregarding irrelevant or misleading information. This two-step approach provides an efficient and effective solution for addressing spurious correlations in image classification.

The combination of accurate detection scores, automated sorting based on the spuriosity, and targeted retraining of the last layer ensures improved reliability and robustness in object classification tasks. This technique also enables the determination of the degree of spuriosity in each image, providing valuable insights into the reliability and accuracy of the detected objects. Additionally, the novel sorting technique eliminates human supervision, which is particularly advantageous when working with large datasets.

\section{Experiments}

To demonstrate the efficacy of our ranking framework, we conduct several experiments which will be discussed in the following sections.



\begin{figure}[t]
    \begin{minipage}[t]{0.5\linewidth}        \includegraphics[width=\linewidth]{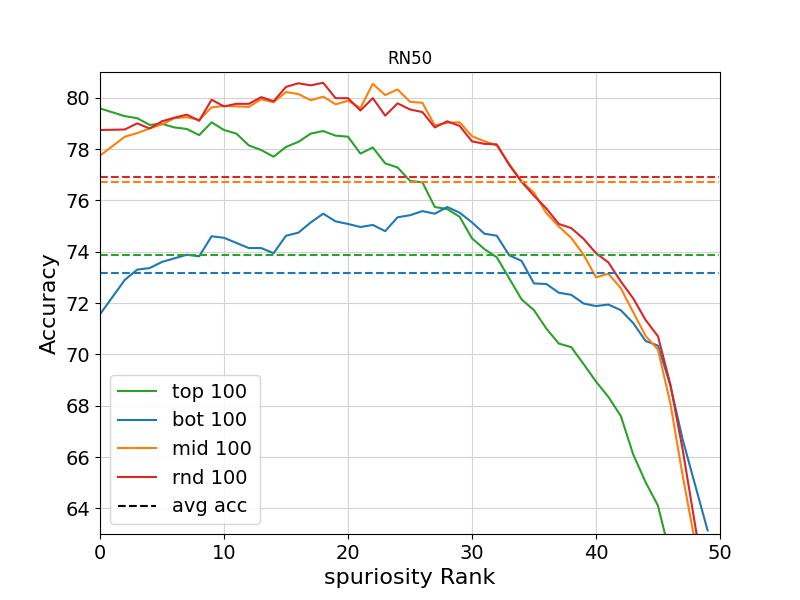}
    \end{minipage}%
    \hfill
    \begin{minipage}[t]{0.5\linewidth}
        \includegraphics[width=\linewidth]{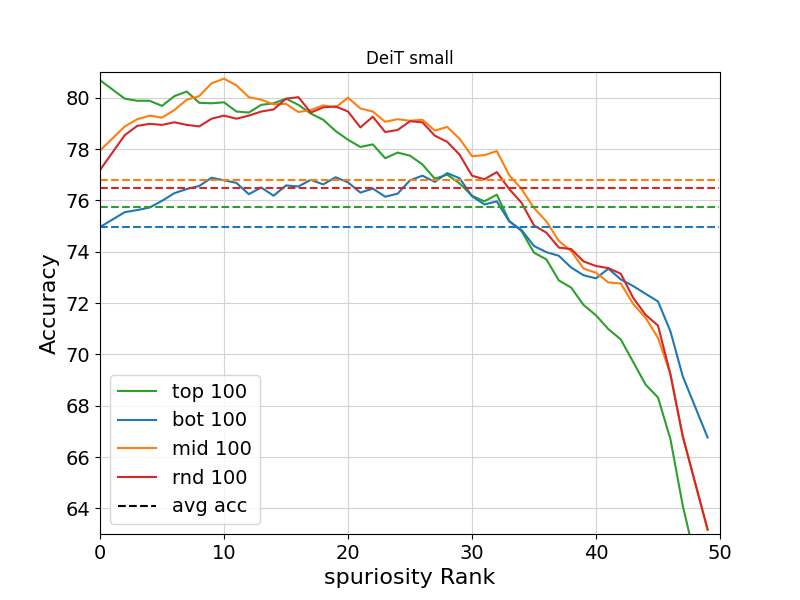}
    \end{minipage} 
    \vskip -0.05in
    \caption{Impact of Training Subsets on Classification Accuracy for Validation Samples with Varying spuriosity Ranks. The dashed horizontal lines depict the average accuracy levels achieved by each model.}
    \label{fig:both}
    \vskip -0.05in
\end{figure}

\subsection{Effect of spuriosity Rankings
} \label{3.1}

In this section, we utilize two pre-trained models  on the ImageNet-1k (ResNet-50 or DeiT-small) and try to re-train only their last layer as a linear classification head on a subset from the same data from scratch. This process is similar to the method used in \cite{lastlayer}.
The training data subsets are selected based on our proposed approach for spuriosity rankings. More specifically, we consider four subsets each containing k samples of each class: \textit{bot k} (lowest spuriosity rank), \textit{mid k} (closest to median rank), \textit{top k} (highest spuriosity rank), and \textit{rnd k} (randomly selected samples). By incorporating these various subsets of data, we aim to capture the impact of our proposed method for ranking images in the last layer re-training method.
For assessing the models' performance, each model is evaluated on varying spuriosity level test data. For this purpose, we chose the image with the i-th highest spuriosity rank from each class in the test data, resulting in a set of images with varying spurious rankings. 


The evaluation results for k=100 are presented in \cref{fig:both}, indicating that the model trained on \textit{top k} data performs better on low-ranked ones, whereas the model trained on \textit{bot k} data is more resilient to the accuracy change when assessed on data with varying spurious rankings. It is worth noting that our method achieves similar results to \cite{moayeri2022spuriosity}, but ranks images without human supervision. The results for k=50 and k=200 could be found in the appendix.

\begin{figure}[t]
    \begin{minipage}[t]{0.5\linewidth}        \includegraphics[width=\linewidth]{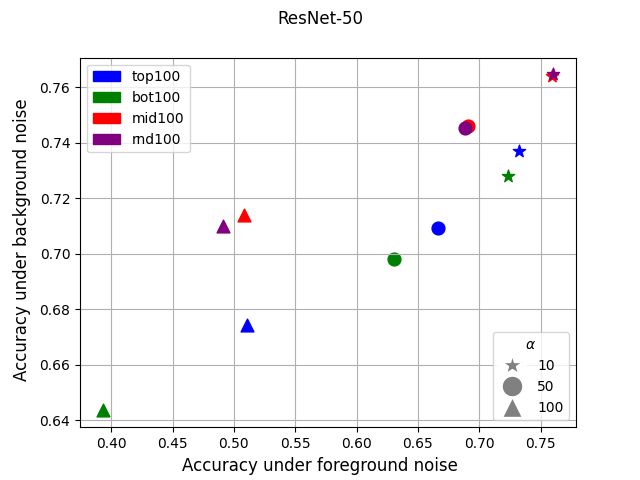}
    \end{minipage}%
    \hfill
    \begin{minipage}[t]{0.5\linewidth}
        \includegraphics[width=\linewidth]{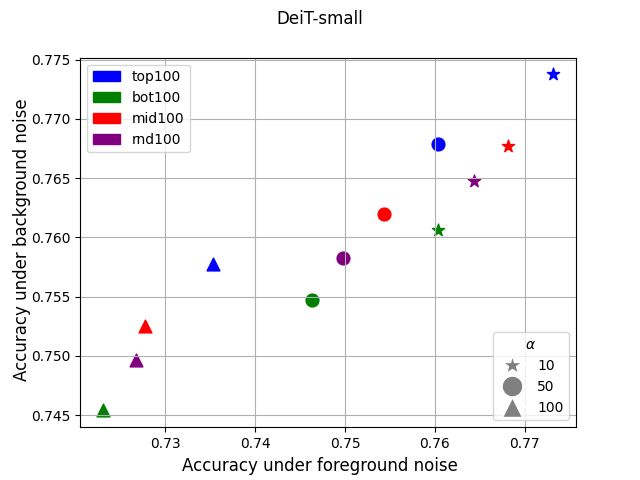}

    \end{minipage} 
    \vskip -0.05in
    \caption{Experimental results show \textit{top 100} models are more robust than \textit{bot 100} under foreground and background noise for both ResNet and VIT models, with \textit{top 100} VIT outperforming \textit{rnd 100} and \textit{mid 100} models. Results suggest \textit{top 100} approach enhances ResNet's robustness on high-level noisy foreground samples.}
    \label{fig:both2}
    \vskip -0.05in
\end{figure}

\subsection{Foreground and Background Noise Sensitivity Analysis: Comparing Model Robustness}
In this section, we present a sensitivity analysis on the models trained as described in section \ref{3.1} by adding noise to the foreground or background of the images.
First, we define the foreground region of ImageNet images as the area inside the detection rectangle determined by the OWL-ViT model. We create a binary mask for each sample by assigning a value of 1 to pixels within the foreground region and 0 otherwise. To create the noisy samples, a noise tensor $n$ is generated in which the pixel values are independently drawn from 
$\mathcal{N}(0, 1)$. Afterward, the noisy foreground sample $\bar{x}_{f g}$ and noisy background sample $\bar{x}_{b g}$ are: $$
\begin{gathered}
\bar{x}_{f g}=x+\alpha \cdot f\left(n \odot m_{x}\right) \\
\bar{x}_{b g}=x+\alpha \cdot f\left(n \odot\left(1-m_{x}\right)\right),
\end{gathered}
$$ where $\odot$ is the hadamard product, $m_{x}$ is the binary mask for sample $x$, $f($.$)$ is the noise $\ell_2$ normalization function, and $\alpha$ is a scalar parameter controlling noise level. $\ell_2$ normalization is applied to ensure that the magnitude of noise in the foreground and background remains consistent, thereby enabling fair comparisons between them. 
We consider three different levels of noise $\alpha \in\{10,100,250\}$ for evaluating the models' sensitivity. \cref{fig:both2} shows the results of our experiment, where the \textit{top 100} models consistently exhibit greater robustness than the \textit{bot 100} models under foreground and background noise for both ResNet and VIT models. The \textit{top 100} VIT model outperforms the \textit{rnd 100} and \textit{mid 100} models, while the \textit{mid 100} and \textit{rnd 100} ResNet models demonstrate greater robustness due to their ability to learn more robust features of the foreground region in the presence of low-level noise. The results suggest that the \textit{top 100} approach enables ResNet to achieve greater robustness on high-level noisy foreground samples, which confirms the effectiveness of our approach.

\subsection{Model Performance on OOD Data}

\begin{figure}[t]
\begin{center}
\vskip -0.12in
\centerline{\includegraphics[scale=0.25]{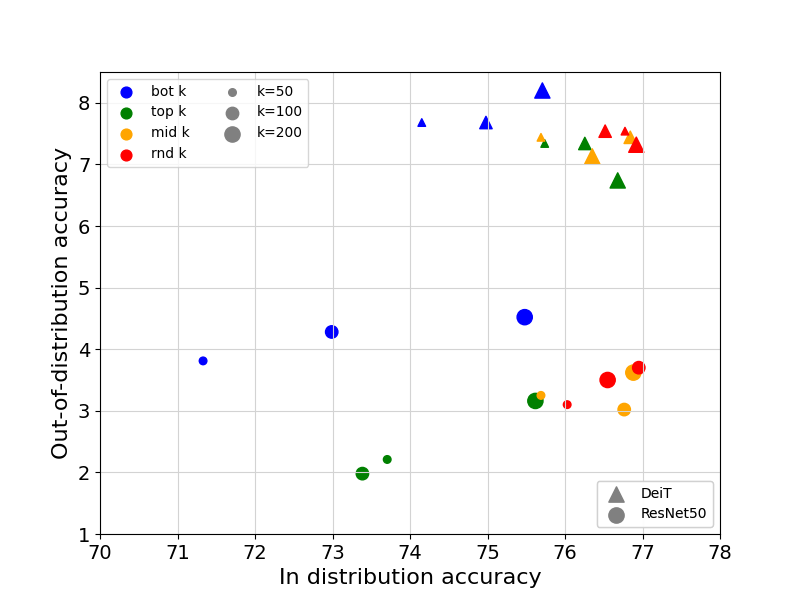}}
\vskip -0.1in
\caption{The performance comparison of different models on OOD dataset (ImageNet-A) and in-distribution setting. VIT models outperform others on OOD data, while the \textit{bot k} subset models achieve superior results on ImageNet-A; \textit{mid k} and \textit{rnd k} subset models demonstrate strong performance across both in-distribution and OOD scenarios.}
\label{ood}
\end{center}
\vskip -0.1in
\end{figure}

In this experiment, we aim to evaluate the performance of various models on OOD datasets, specifically ImageNet-A. The models used are the same as those trained in section \ref{3.1}, and their accuracy is assessed on both in-distribution and OOD data. The results are presented in \cref{ood}, indicating that VIT models outperformed convolutional models on OOD dataset. Moreover, models trained on \textit{bot k} achieved higher accuracy on ImageNet-A, possibly due to the similarity in image style between ImageNet-A and the training data for \textit{bot k} models. This suggests that the \textit{bot k} models were able to effectively capture the underlying patterns and characteristics of the ImageNet-A images, resulting in improved performance on this specific OOD dataset. Additionally, models trained on \textit{mid k} and \textit{rnd k}, which include different image styles, demonstrate superior overall performance compared to other models in both in-distribution and OOD settings.

\section{Conclusion}
In conclusion, our study introduced a novel method for ranking images based on their spuriosity within classes without human supervision. Through comprehensive experiments, we examined the impact of our approach on model training, assessed model sensitivity to foreground and background noise, and evaluated model performance on OOD data. The results demonstrate the efficacy of our method in enhancing model's robustness and performance in the presence of spurious cues. By eliminating the need for human supervision, our approach presents a promising solution for tackling spurious features and strengthening the robustness of deep neural networks in practical applications.

\bibliography{example_paper}
\bibliographystyle{icml2023}

\newpage
\appendix
\onecolumn
\section{Appendix}

\subsection{Experimental Analysis of Training Subsets and spuriosity Ranking on Model Performance}

We conducted comprehensive experiments using two specific values of k, k=50 and k=200, to further investigate the impact of training subsets on model's performance. The models were trained using the same method described in section \ref{3.1}, on the proportion of the data sorted based on spuriosity ranking. For k=50, the evaluation results revealed that the model trained on the \textit{top 50} subset exhibited superior performance on low-ranking data, showing its ability to effectively capture and generalize patterns from such samples. On the other hand, the model trained on the \textit{bot 50} subset demonstrated increased resilience to accuracy changes when assessed on data with varying spurious rankings. This indicates that the model trained on the \textit{bot 50} subset is more robust and less influenced by spurious cues present in the test data.

Similarly, for k=200, we observed a similar trend. The model trained on the \textit{top 200} subset displayed better performance on low-ranking data, while the model trained on the \textit{bot 200} subset exhibited higher resilience to accuracy changes when evaluated on data with varying spurious rankings.

\begin{figure}
  \centering
  \includegraphics[width=\linewidth]{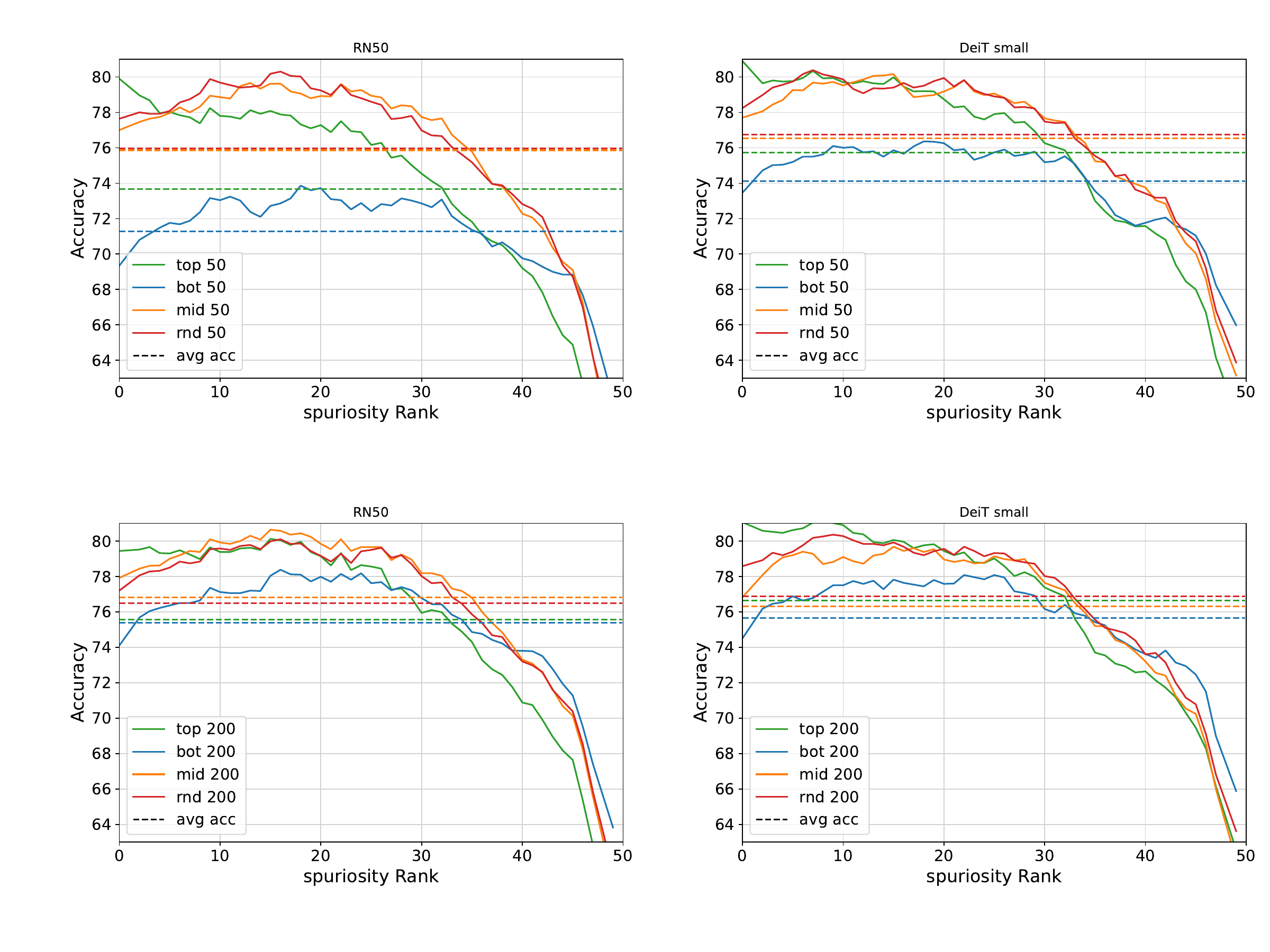} 
  \caption{Impact of training subsets (k=50, 200) on classification accuracy for validation samples with Varying spuriosity ranks. The plot illustrates how different subsets of training data, each containing 50 and 200 samples per class respectively, influence the accuracy of classification. The dashed horizontal lines represent the average accuracy levels achieved by each model
    on different data subsets.}
  \label{fig:both3}
\end{figure}

These findings highlight the impact of different training data subsets and spuriosity rankings on the model's performance. The models trained on subsets containing low-ranking samples prioritize accuracy on those data points, while the models trained on subsets with high-ranking samples are more robust to changes in spurious correlations. These insights contribute to a deeper understanding of the relationship between training data subsets, spuriosity rankings, and model performance.
The detailed visualizations and analysis of the results for k=50 and k=200 are presented in \cref{fig:both3}.

\subsection{
Further Illustrations of spuriosity Rankings}

\begin{figure*}[t]
  \centering
  \includegraphics[scale=0.45]{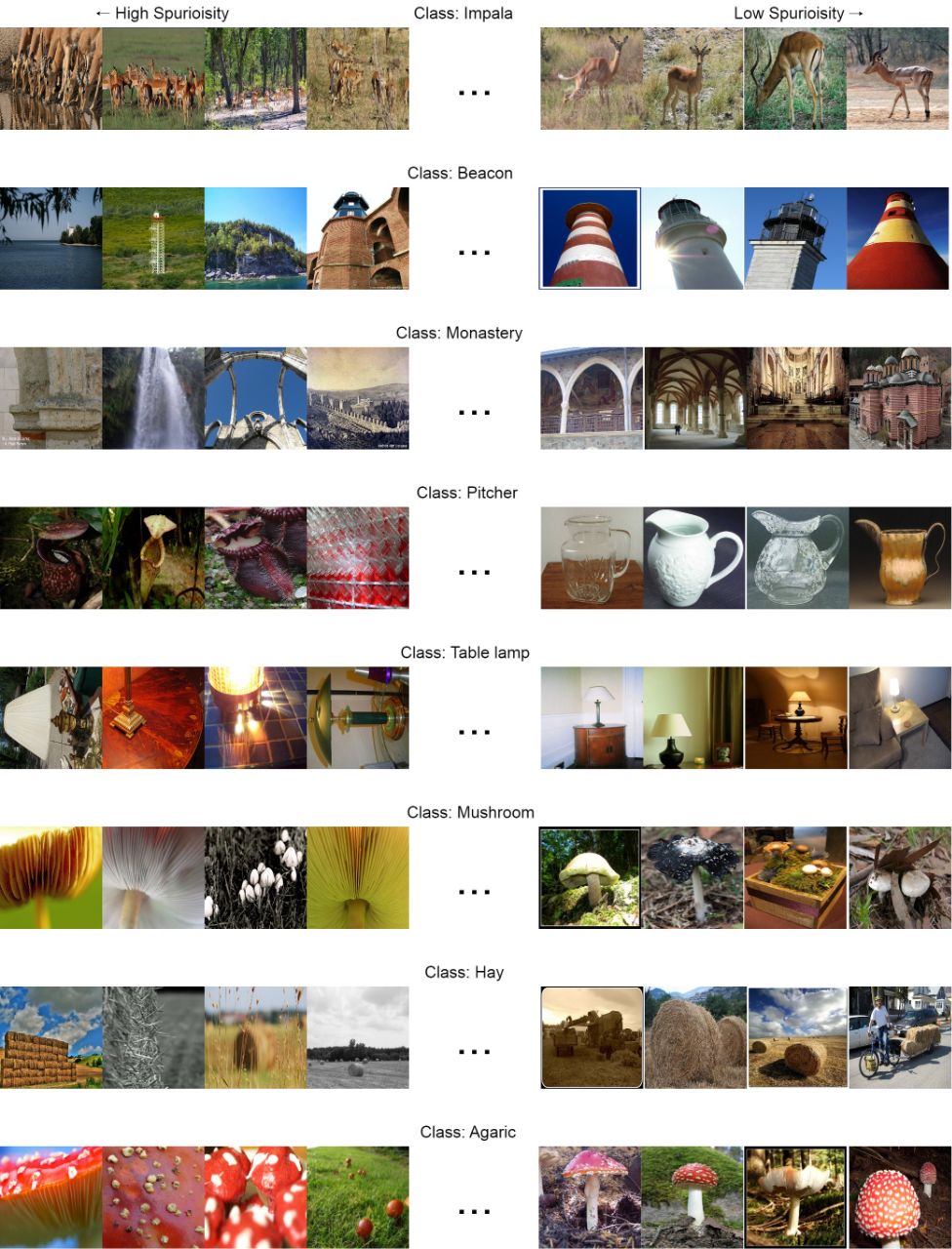} 
  \caption{Examples of Spuriosity Rankings: Illustrating object detectability and spurious cues. The images on the left side demonstrate high spuriosity, showing cases where spurious cues are present, making the object difficult to distinguish. Conversely, the images on the right side exhibit low spurious cues, displaying clear images of the object.}
  \label{fig8} 
\end{figure*}

In this part, we present additional examples that highlight the concept of spuriosity rankings, their implications for object detectability, and the presence of spurious cues for some random classes. A wide range of images is shown in \ref{fig8} ranging from clear images with low spuriosity to hard-to-distinguish cases with high spuriosity.

\end{document}